\pdfoutput=1
\documentclass[11pt]{article}

\usepackage[preprint]{acl}
\usepackage{times}
\usepackage{latexsym}
\usepackage[T1]{fontenc}
\usepackage[utf8]{inputenc}
\usepackage{microtype}
\usepackage{inconsolata}
\usepackage{graphicx}
\usepackage{hyperref}
\usepackage{url}
\usepackage{graphicx}
\usepackage{booktabs}
\usepackage{multirow}
\usepackage{makecell}
\usepackage{lipsum}
\usepackage{amsmath}
\usepackage{amssymb}
\usepackage{arydshln}
\usepackage{tabularx}
\usepackage{algorithm}
\usepackage{algorithmic}
\usepackage{tcolorbox}
\usepackage{colortbl}
\usepackage{xcolor}
\usepackage{soul}
\usepackage{xspace}
\definecolor{teacher}{HTML}{0E8088}
\definecolor{assistant}{HTML}{B46504}
\definecolor{classmate1}{HTML}{AE4132}
\definecolor{classmate2}{HTML}{56517E}
\definecolor{user}{HTML}{56517E}
\definecolor{student_direct}{HTML}{DFDDF2}
\newcommand\modelname{SimClass\xspace}

\title{Simulating Classroom Education with LLM-Empowered Agents}

\author{Zheyuan Zhang$^\spadesuit$\thanks{\quad Equal contribution}, Daniel Zhang-Li$^\spadesuit$\footnotemark[1], Jifan Yu$^\heartsuit$, Linlu Gong$^\spadesuit$, \\ \textbf{Jinchang Zhou}$^\spadesuit$, \textbf{Zhanxin Hao}$^\heartsuit$, \textbf{Jianxiao Jiang}$^\heartsuit$, \textbf{Jie Cao}$^\heartsuit$, \\ \textbf{Huiqin Liu}$^\heartsuit$, 
\textbf{Zhiyuan Liu}$^\spadesuit$, \textbf{Lei Hou}$^\spadesuit$, \textbf{Juanzi Li}$^\spadesuit$ \\
        $^\spadesuit$Department of Computer Science and Technology; $^\heartsuit$Institute of Education \\
        Tsinghua University, Beijing, 100084, China \\ \texttt{zheyuan-22@mails.tsinghua.edu.cn} \quad
        \texttt{lijuanzi@tsinghua.edu.cn}}

\begin{document}
\maketitle
\begin{abstract}
Large language models (LLMs) have been applied across various intelligent educational tasks to assist teaching. While preliminary studies have focused on task-specific, independent LLM-empowered agents, the potential of LLMs within a multi-agent collaborative framework for classroom simulation with real user participation remains unexplored. In this work, we propose \modelname, a multi-agent classroom simulation teaching framework. We recognize representative class roles and introduce a novel class control mechanism for automatic classroom teaching, and conduct user experiments in two real-world courses. Using the Flanders Interactive Analysis System and Community of Inquiry theoretical frameworks from educational analysis, we demonstrate that LLMs can simulate a dynamic learning environment for users with active teacher-student and student-student interactions. We also observe group behaviors among agents in \modelname, where agents collaborate to create enlivening interactions in classrooms to improve user learning process. We hope this work pioneers the application of LLM-empowered multi-agent systems in virtual classroom teaching.
\end{abstract}

\section{Introduction}

The pursuit of utilizing artificial intelligence to provide immediate and customized teaching for students origins from the era of Intelligent Tutoring Systems (ITS)~\cite{nwana1990intelligent}. Following this enthusiasm, from personalized educational recommendation systems~\cite{liu2019exploiting} to teaching assistants~\cite{tu2023littlemu, khanmigo2024} and even LLM-driven AI teacher~\cite{markel2023gpteach, yue2024mathvc}, researchers have conducted enormous technological explorations and achieved impressive performance in specific educational tasks. 

As technology advances, intense discussions have also emerged around this topic concerning methodologies~\cite{extance2023chatgpt,yue2024mathvc}. One of the most central directions is how to fully leverage the capabilities of large models to \textbf{simulate real classrooms with multiple agents for automated teaching}. 
From an educational perspective, this approach allows large models to move beyond their instrumental use and delve deeper into educational paradigms~\cite{lave1996teaching,opara2023chatgpt}. From a technical standpoint, multi-agent collaboration technologies~\cite{qian2023communicative} could further stimulate the latent knowledge of large models in education, leading to the emergence of richer capabilities~\cite{li2024camel,aher2023using}. 

However, several fundamental research questions for LLM-empowered multi-agent systems with real user participation remain: (1) \textit{Simulation Performance}: How well can a multi-agent classroom simulate real-time teacher-student interactions? (2) \textit{Learning Experience}: Can students in such environment experience a high sense of presence and learn effectively?
(3) \textit{Group Behavior Observation}: What behaviors may arise spontaneously in multi-agent scenarios?

\begin{figure*}[ht]
  \includegraphics[width=0.85\linewidth]{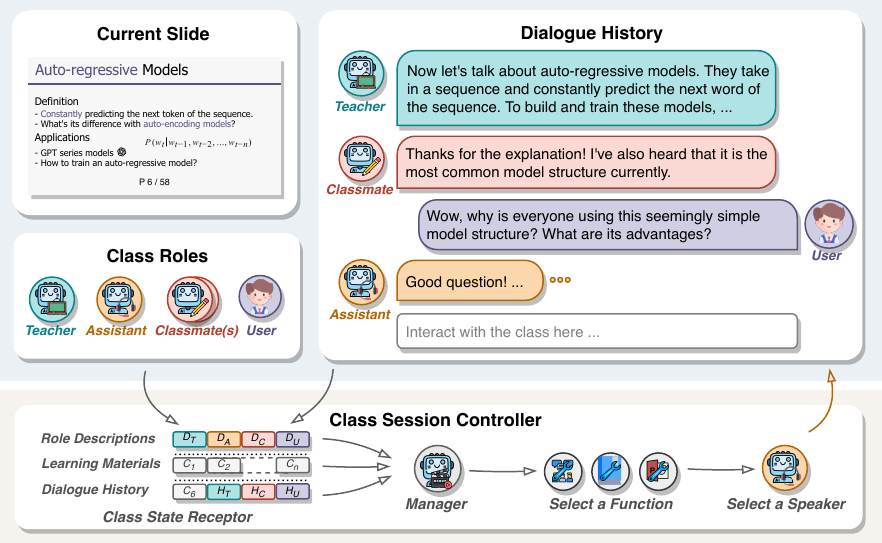}
  \centering
  \caption{An overview of the \modelname framework. Note that the upper portion of the framework is visible to student users, while the lower portion is hidden from them. In the classroom, users can view the current slide, configure class roles, and engage in real-time conversations with the agents.}
  \label{fig:simclass}
\end{figure*}

In response to the research questions above, we present \textbf{\modelname}, a multi-agent classroom simulation framework, and conduct real-world experiments and analysis on it. For better simulation, we identify representative class roles and design a novel class control mechanism (Figure~\ref{fig:simclass}).
We deploy two different courses with prepared slides and teaching scripts as the foundation. We conduct online experiments with over 400 students, who participated in the courses and interacted with the system, with all the behavioral data carefully recorded. Additionally, we constructed ablation systems and invited another 48 students for further experiments. Our research addresses the following parts: (1) We apply the Flanders Interaction Analysis System~\cite{amatari2015instructional} to evaluate the interactions within the \modelname and examine the teaching style of the agents. (2) We analyze the learning outcomes and educational experiences of these users, using Community of Inquiry theory~\cite{garrison2007researching}. (3) Lastly, we identify group behaviors of agents for qualitative analysis.

The experimental results demonstrate the effectiveness of the class roles and control mechanism design in the following aspects: (1) \textbf{Performance}: \modelname fosters a vivid learning environments with lively teacher-student and student-student interactions; (2) \textbf{Experience}: Students retain the knowledge gained in the \modelname, with increased interactions contributing to improved learning outcomes. The presence of multiple classroom agents enhances user engagement and strengthens their sense of presence; (3) \textbf{Behavior}: The control mechanism elicits group behaviors in the multi-agent classroom, including collaborative teaching, discussions, emotional company and discipline management. In summary, the LLM-based multi-agent system shows great potential for simulating classroom for educational purposes. We hope our work serves as a pioneering effort in this area. The dataset of classroom interactions will be released soon for both education and AI researchers.

\section{Related Work}

\subsection{LLMs for Human Simulation}

Recently, Large Language Models (LLMs) have achieved remarkable breakthroughs in various natural language processing (NLP) tasks~\cite{brown2020language, achiam2023gpt, touvron2023llama, reid2024gemini}. The intelligence they demonstrated opened up opportunities and possibilities for applications in many other scenarios~\cite{bubeck2023sparks, yang2023dawn}. As LLMs encode many human-like behaviors in their training data, an increasing number of researchers are utilizing LLMs for human scenario simulation, investigating the model's capabilities for decision and actions as LLM-Empowered Agents in many fields, such as social and psychological research~\cite{aher2023using, park2023generative, li2024camel, gao2023s, li2023large, zhang2023exploring}, software development~\cite{qian2023communicative, hong2023metagpt}, chemical and medicine~\cite{li2024agent, m2024augmenting}, and games~\cite{wang2023voyager}. Novel collaboration techniques are explored to enhance the cooperation and performance of multi-agent systems~\cite{cheng2024cooper, wu2023autogen}. These works offer technical possibilities for multi-agent education and inspire curiosity about potential emergent phenomena.

\subsection{LLMs for Education}

With the eminent linguistic capabilities, explanatory skills, and parameterized knowledge of LLMs, numerous studies have explored applying LLMs to education services. In addition to applying large models to downstream tasks in the education~\cite{hu2024teaching, li2024explainable, jeon2023large}, many researchers are applying these models to replace certain classroom aspects, such as playing students to train teachers~\cite{LeeGenerativeAF, markel2023gpteach} or playing instructors to teach students~\cite{tu2023littlemu, sonkar-etal-2023-class, khanmigo2024, chen2023empowering}. \citet{yue2024mathvc} explored the use of multiple student agents to assist students in discussion, though they haven't involved real users. Existing work has examined various facets of interactions between LLMs and humans in educational settings.

\section{\modelname}
\label{sec:simclass}

\subsection{Overview} 

The design principles for constructing this immersive simulated classroom originate from the following two concerns: (1) How to ensure that the classroom covers the core teaching behaviors? (2) How to maintain the entirety of the interaction within the natural flow of the classroom process? 

For the former concern, we categorize classroom interaction behaviors based on widely accepted pedagogy principles~\cite{schwanke1981classroom}: \textit{Teaching and Initiation (TI)}, the teacher's teaching and the students' feedback or ideas; \textit{In-depth Discussion (ID)}, alignment, discussion, and multiple Q\&A to help students construct understanding of concepts; \textit{Emotional Companionship (EC)}, encouraging students to learn, creating a positive learning atmosphere, and providing emotional support; and \textit{Classroom Management (CM)}, maintaining discipline, organizing disruptive behaviors, and guiding the classroom content. 
Given that these behaviors are realized through the varied \textbf{Class Roles} (denoted as $\widehat{ \mathcal{R}}= \left\{r_{i} \right\}_{1}^{\left| \widehat{ \mathcal{R}} \right|}$, where each $r_{i}$ denotes a certain role), it is essential to ensure the \textit{diversity} and \textit{coverage} of proposed agents within the classroom.

For the latter concern, we need to ensure that the interactions among multiple agents within the system are finely and rhythmically controlled within the course content. As shown in Figure~\ref{fig:simclass}, given the Class Roles and Learning Materials (denoted as $C = \left [ c_1,...,c_t \right ]$, where each teaching script $c_t$ is organized by order), we propose a novel \textbf{Session Controller} to manage the course interaction flow based on class status and the help of a core manager agent~\cite{wu2023autogen}.

Based on these principles, we construct multiple class roles, implement class control, and ultimately derive the simulated classroom process. Their prompts are shown in Appendix.

\subsection{Class Role Agentization}

The teaching process is presented as an informative, multi-round, and task-oriented communication~\cite{lave1996teaching}. However, simply exchanging responses of LLMs inevitably faces significant challenges including role flipping, instruction repeating, and fake replies~\cite{qian2023communicative}. Consequently, following the classroom behaviors outlined previously, we define two types of agents: \textit{Teaching Agents} and \textit{Classmate Agents}. Each agent $\mathbf{a}_{i} \in \mathcal{A}$ is facilitated through prompting LLMs and associated with one or more class roles, denoted as:
\begin{equation}
     \mathcal{A} = \rho \left ( LLM, \mathsf{P}_{A} \right ), \mathcal{A} \Leftrightarrow \widehat{\mathcal{R}}
\end{equation}
where $\rho$ is the role customization operation, $\mathsf{P}_{A}$ is the system prompt with agent description. All roles designs and corresponding prompts were crafted with input from experienced teaching practitioners.  Relevant technologies, such as question generation~\cite{kurdi2020systematic} and retrieval-augmented generation~\cite{lewis2020retrieval}, can also be integrated into the construction of class roles.

\paragraph{Teaching Agents} The teacher and the teaching assistant are the authoritative party responsible for imparting knowledge in the classroom, encompassing most teaching behaviors. The acronyms in parentheses represent the roles that the agent needs to accomplish in a classroom environment.
 
\textbf{\textcolor{teacher}{\textit{Teacher Agent (TI, ID, EC, CM)}} }: Given the teaching scripts $C$, its task is to persuasively display material $c_i$ to students or answer questions based on the classroom historical discussions $H$.

\textbf{\textcolor{assistant}{\textit{Assistant Agent (ID, EC, CM)}}}: Given the classroom history $H$, the assistant is responsible to supplement teaching information, participate in discussion, maintain the discipline and continuity of the class, and enhance student learning efficiency. 

\paragraph{Classmate Agents} This type of agents are incorporated in addition to the teaching agents with distinct personality traits to perform peer student roles. In this paper, we initialize $4$ typical classmates, while users can also freely customize more interesting classmate agents on the platform.

\textbf{\textcolor{classmate1}{\textit{Class Clown (TI, EC, CM)}}}: This agent is designed to initiate ideas, enliven the atmosphere, help the user as a peer, and help the teachers to guide the class direction when the user is distracted.

\textbf{\textcolor{classmate1}{\textit{Deep Thinker (TI, ID)}}}: This agent aims to do deep thinking and raise topics that challenge the knowledge of the classroom.

\textbf{\textcolor{classmate1}{\textit{Note Taker (TI, CM)}}}: This agent loves to summarize and share notes for classroom content, helping everyone to organize their thoughts.

\textbf{\textcolor{classmate1}{\textit{Inquisitive Mind (TI, EC)}}}: This agent frequently poses questions about lectures, which stimulates others' thinking and discussion.

\subsection{Classroom Session Controller}

Unlike multi-agent systems with Standardized Operating Procedures (SOPs)~\cite{qian2023communicative, hong2023metagpt}, the classroom scenario is a dynamic group chat without a strict workflow, requiring agents to determine appropriate speaking times on the fly. Therefore, we implement a Session Controller that observes, decides, and directs agent behavior based on the current Class State. It comprises three modules: Class State Receptor, Function Executor, and Manager Agent.

\textbf{Class State Receptor.} Let the classroom dialogue history until time $t$ denote as $H_t = \bigcup (u_i^{\mathbf{a}_j})^{t}$, where $u_i$ is the utterance posted by agent $\mathbf{a}_j$ or user (denoted as $\mathbf{a}_{u}$). The class state $S_t$ is composed as:

\begin{small}
    \begin{equation}
    \mathcal{S}_t = \left\{C_t, H_t | \widehat{\mathcal{R}}   \right\} 
\end{equation}
\end{small}

where $C_t \subseteq C $ is composed of the learning materials that have been taught until $t$.

\textbf{Functions.} We design and divide the actions in the classroom into a functional hierarchy with two major categories. Tutoring functions $f_X$ can only be performed by teacher agent $\mathbf{a}_0$, such as teaching by displaying scripts and going to the next material page $c_{i+1}$. Interacting functions $f_Y$ can be performed by each agent $\mathbf{a}_j \in \mathcal{A}$. According to the context, the interaction will emerge as diverse classroom activities, which are discussed in subsequent experiments. These functions are pluggable, allowing the addition of newly defined functions for different agents, such as displaying exercises. 

\begin{small}
\begin{equation}
f = \left\{\begin{matrix} f_X
\left\{\begin{matrix}
f_{0}(c_i, \mathbf{a}_{0}), & \text{Teaching.} \\ f_{1}(c_{i+1}, \mathbf{a}_{0}), & \text{Next Page.} \\ ... & \text{...}
\end{matrix}\right. \\ 
f_{Y} \left\{\begin{matrix} f_{n}(c_i, \mathbf{a}_{j}, H_t), &     \text{Interaction.} \\ ... & \text{...}
\end{matrix}\right.
\end{matrix}\right.
\end{equation}
\end{small}

\textbf{Manager Agent.} Following AutoGen~\cite{wu2023autogen} and MathVC~\cite{yue2024mathvc}, we design a hidden and meta agent to regulate the speakers. This agent receives the current class state $\mathcal{S}_t$, observes and understands the class process, and decides the next action to be executed. The task $\mathcal{L}$ of Manager Agent can be defined as:
\begin{equation}
    \mathcal{L}: \mathcal{S}_t \rightarrow \left ( \mathbf{a}_t, f_t \right ) | \mathbf{a}_t \in \mathcal{A}, f_t \Leftarrow  f
\end{equation}
where $f_t$ is a certain kind of function, and the action will be executed and refresh the whole class into the next state. Specifically, the system will wait for a time window $\tau $ after an action is performed. If the user speaks or the waiting period ends, it will trigger the manager agent to make a new decision.

\subsection{Classroom Demonstration}

After introducing the necessary components of the \modelname, we demonstrate a complete class process: (1) \textbf{Initialization}. At the start, the first function executes, displaying the initial script and slides. Users can begin interacting, and the manager agent takes control of the class flow; (2) \textbf{Tutoring and Interaction}: the manager agent will continuously observe and control the class based on the states, selecting appropriate functions and speakers, and coordinates agent collaboration. As shown in Figure~\ref{fig:simclass}, when a user asks about the content, the classroom interaction flow may involve the assistant responding, the teacher adding details, and sometimes the classmate agents raising relevant topics; (3) \textbf{Ending}. After all the learning materials are taught and the final discussion ends, the classroom will close and provide quizzes to users. 


\section{Experiments}

We focus on three research questions to evaluate and understand \modelname as a multi-agent learning environment: (1) What is the performance of \modelname? (2) What are the impacts of various interaction types within \modelname (e.g. the roles of student agents)? (3) How do agents behave in \modelname? To address these questions, we deploy \modelname online and invite a group of university students to use the system. With the Institutional Review Board (IRB) approval from our institution, we design quizzes and collected interaction data for further analysis. Additionally, we develop ablation systems to investigate the influence of class roles and interactions in the environment.

\subsection{Experimental Setup}

\textbf{Courses and Materials.} 
We conduct experiments with two courses. The first, \textit{TAGI, Towards Artificial General Intelligence}, covers AI development and language models across six meticulously designed chapters. The second, \textit{HSU, How to Study at University}, focuses on academic skills, stress management, comminication, and self-fulfillment, spanning seven well-structured chapters. While both courses contain structured slides and teaching scripts, TAGI emphasizes knowledge acquisition, and HSU aims at skill development.

\textbf{Systems.}
We use GLM-4~\cite{glm2024chatglm} as the backbone LLM for both Class Roles and Manager Agent due to cost and concurrency constraints in the online system. To explore the impact of class roles, we deploy three ablation systems with the model replaced by GPT-4V. The first replicates the original system, the second removes classmate agents (retaining only the teacher and assistant), and the third disables both classmate agents and user input, with the teacher conducting uninterrupted lectures LLMs can effectively deliver courses without modifying agent prompts.

\textbf{Participants.}
We recruited over 400 university students from various majors for the online learning system, with 118 completing all of the chapters (77 in TAGI, 41 in HSU). An additional 48 students participated in the ablation study for only the first course chapter. To ensure data quality of ablation systems, students took a brief test after the course, assessing whether they used the system and remember some of the basic concepts covered, and data from those scoring below 50\% was excluded. Participants were informed that course content is AI-generated, and their interaction data would be only used for research. All participants were compensated above the national average hourly wage.

\textbf{Data Collection.} The data collected for the online and the ablation systems emphasizes different aspects to address specific research questions:

(1) Online System: We meticulously recorded all user interactions for interaction analysis (Section~\ref{sec:interaction}). To evaluate students' learning outcomes, we invited educational practitioners to design quizzes after each chapter and a final exam for the knowledge acquisition course, TAGI, to assess knowledge retention (Section~\ref{sec:outcome}). For the practical course, HSU, we employed a self-reported method, where students wrote self-summaries.  

(2) Ablation systems: We examined how class roles affect learning by tracking interactions and developing a brief survey based on the widely recognized Community of Inquiry (CoI) theory~\cite{garrison1999critical}. The survey measures three elements: \textit{Cognitive Presence}, the degree to which learners are able to construct and confirm meaning through sustained reflection and interaction; \textit{Teaching Presence}, the extent to which the class is focused, designed, and planned with specific learning objectives; and \textit{Social Presence}, the ability of learners to project themselves socially and emotionally within a group~\cite{garrison2007researching}. Following prior research~\cite{yu2022xdai, tu2023littlemu}, students rated the system on a [0,1,2] scale according to detailed guidelines, with higher scores indicating better performance. Survey questions and guidelines are detailed in Appendix~\ref{apd:survey}.

\subsection{Online System Results}

We demonstrate the performance of \modelname in the online system by analyzing student interactions and their learning outcomes.

\begin{figure*}[t]
  \includegraphics[width=0.85\linewidth]{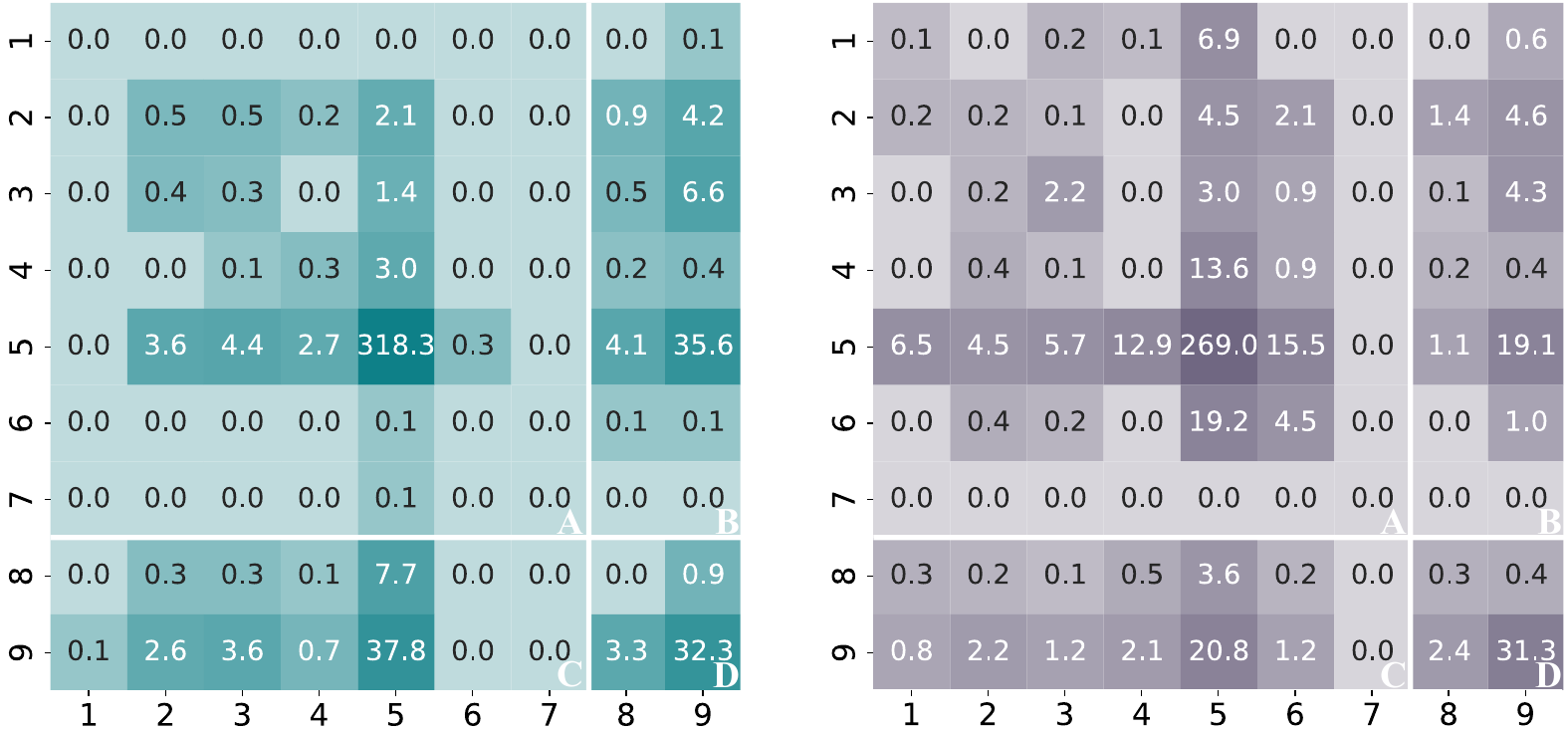}
  \centering
  \caption{The FIAS matrix sum of users in TAGI (left) and HSU (right). Numbers \(1\text{--}9\) represent the corresponding categories. \( N \) in location \(( x, y )\) means that there are \( N \) transitions from \( x \) to \( y \) in the classroom. The matrix is divided into four parts based on the type of interaction between actors.}
  \label{fig:matrix}
\end{figure*}

\subsubsection{Interaction Analysis}
\label{sec:interaction}

To understand the dynamics of \modelname as a multi-agent classroom system, we encode classroom activities into quantitative behaviors. We employ the Flanders Interaction Analysis System (FIAS)~\cite{amatari2015instructional}, a widely adopted tool for analyzing verbal behaviors in traditional classrooms. We adapt the method to our simulated classroom system, where interactions occur in natural language, to investigate the teaching patterns of \modelname.

\begin{table}[ht]
\centering
\small
\begin{tabular}{l l l}
\toprule
\textbf{Speaker} & \textbf{Type} & \textbf{Action} \\ 
\midrule
\multirow{7}{*}{Teacher} & \multirow{4}{*}{\begin{tabular}[c]{@{}l@{}}Indirect \\ Influence\\ (Response)\end{tabular}} & 1. Accept Feelings          \\ \cmidrule{3-3} 
                         &       & 2. Praises or Encourages               \\ \cmidrule{3-3} 
                         &       & 3. Accept Ideas    \\ \cmidrule{3-3} 
                         &   & 4. Ask Questions      \\ \cmidrule{2-3} 
                         & \multirow{3}{*}{\begin{tabular}[c]{@{}l@{}}Direct\\ Influence\\ (Initiation)\end{tabular}}  & 5. Lecturing                           \\ \cmidrule{3-3} 
                         &          & 6. Giving Direction    \\ \cmidrule{3-3} 
                         &             & 7. Criticizing \\ 
                         \midrule
\multirow{2}{*}{Student} & Response                                                                                    & 8. Response                            \\ \cmidrule{2-3} 
                         & Initiation      & 9. Initiation          \\ 
\bottomrule
\end{tabular}
\caption{The categories of FIAS.}
\label{tab:fias}
\end{table}

\textbf{Encoding the Interactions.} As shown in Table~\ref{tab:fias}, the FIAS categorizes interactions into nine distinct types: seven for teachers and two for students (we exclude the Silence category from the original FIAS, as it is not applicable to \modelname due to the real-time responses of models and the difficulty of defining or detecting silence in online systems). Labels \(1\text{--}4\) represent Indirect Influence from the teacher, while labels \(5\text{--}7\) indicate Direct Influence.
For the classroom history of each student, we prompt GPT-4 to label interactions according to the nine categories, and assess the quality of labeling in Appendix~\ref{apd:examination}. The classroom interactions are encoded as sequences, and the two-step transitions of classroom activities are recorded in a \(9 \times 9\) matrix \(\mathcal{M} \in \mathbb{N}^{9 \times 9}\). To interpret the Matrix and observe features in \modelname, we report the following frequently used metrics:

\textit{Teacher Talk (TT) and Student Talk (ST).} TT and ST represent the proportions of total tallies in specific categories that indicate the amount of talk from teacher and students. Respectively, TT and ST are calculated using categories \(1\text{--}7\) and \(8\text{--}9\).

\textit{ID Ratio (IDR).} ID Ratio measures the balance between a teacher's indirect and direct methods of teaching in the classroom. It is calculated by dividing the sum of tallies in categories \(1\text{--}4\) (Indirect) by the sum of tallies in categories \(5\text{--}7\) (Direct).

\textit{Student Initiation Ratio (SIR).} SIR evaluates the extent to which students initiate interactions themselves during classroom activities, which measures how much students are actively engaging in the classroom. It is calculated by dividing the tallies in category \(9\) by the total tallies in categories \(8\text{--}9\).


\textbf{Results.} We randomly sampled 10 students each who completed the courses and summed their matrices \(\mathcal{M}_i\) to view the interactions: $\mathcal{M} = \sum_{i=1}^{10} \mathcal{M}_i$. Figure~\ref{fig:matrix} presents the FIAS matrices of \modelname for TAGI and HSU courses, with each divided into four parts based on the type of classroom interaction, labeled as follows:
\textit{A} (top left), Teacher Lecturing: Most teacher actions involve lecturing (Cat. \(5\)), where the teacher primarily delivers lessons and interacts with the class. 
\textit{B} (top right), Student to Teacher: It demonstrates the teacher's responses to students. When students initiate ideas or responses to teachers, the teacher praises (Cat.\(2\)), accepts their ideas (Cat.\(3\)), or continues teaching.
\textit{C} (bottom left), Teacher to Student: It highlights student responses to the teacher, where students frequently ask questions or react to lectures, reflecting the active participation in the classroom.
\textit{D} (bottom right), Student to Student Interactions: This part shows that student-to-student discussions occur periodically. Overall, the classroom exhibits frequent interactions, both between the teacher and students, as well as among the students themselves..

\setul{0.15ex}{0.15ex}
\begin{table}[htbp]
\centering
\small
\begin{tabular}{r | c c c c}
\toprule
\textbf{Course} & \textbf{TT} & \textbf{ST} & \textbf{IDR} & \textbf{SIR} \\
\midrule
TAGI &  0.816 & 0.184 & 0.058 & 0.896 \\
HSU &  0.863 & 0.137 & 0.124 & 0.917 \\
\midrule
ET & 0.771 & 0.229 & 1.473 & 0.121 \\
NT & 0.826 & 0.174 & 0.885 & 0.106 \\
\bottomrule
\end{tabular}
\caption{
Results of the metrics from FIAS, with each number rounded to three decimal places. The ET and NT are short for Expert Human Teacher and Novice Human Teacher reported by~\citet{su15031783}
}
\label{tab:fias_result}
\end{table}

Table~\ref{tab:fias_result} presents the metric results of FIAS, illustrating the teaching style of \modelname. When compared with human classrooms reported by~\cite{su15031783}, the TT and ST ratios are similar, indicating a comparable speaking balance between \modelname and traditional classrooms (with silence removed from both scenarios for a fair comparison). The IDR is relatively low, partly due to the higher proportion of script-based teaching. Meanwhile, the SIR, which reflects the proportion of student-initiated interactions, is relatively high, suggesting a more democratic online learning environment where students feel comfortable asking questions and expressing themselves.

In summary, \modelname fosters a dynamic learning environment with active interactions between teachers and students, as well as among students themselves. Compared to traditional classrooms, students in \modelname are more proactive in initiating discussions and expressing their ideas.

\subsubsection{Learning Outcome Analysis}
\label{sec:outcome}

We assess students' learning outcomes in the online system through quizzes or self reports. In TAGI, all the questions in the quizzes are multiple-choice, with one or more correct answers, requiring all correct answers for full marks. The final exam draws from previous quizzes to evaluate knowledge retention. The average quiz scores (with full mark as 1) is presented in Table~\ref{tab:quiz}. The final exam score aligns with the average quiz scores (0.68), indicating students' consistent retention of the material.

\begin{table}[htbp]
    \centering
    \small
    \begin{tabular}{ccccccc}
        \toprule
        \multicolumn{6}{c}{\textbf{Quiz}} & \textbf{Final} \\ 
        \midrule
        1st & 2nd & 3rd & 4th & 5th & 6th & - \\  
        \midrule
        0.64 & 0.53 & 0.66 & 0.82 & 0.78 & 0.66 & \textbf{0.65} \\  
        \bottomrule
    \end{tabular}
    \caption{Quiz and Final scores. The Final score closely aligns with the average quiz scores (0.68).}
    \label{tab:quiz}
\end{table}

\begin{figure}[htbp]
  \includegraphics[width=\linewidth]{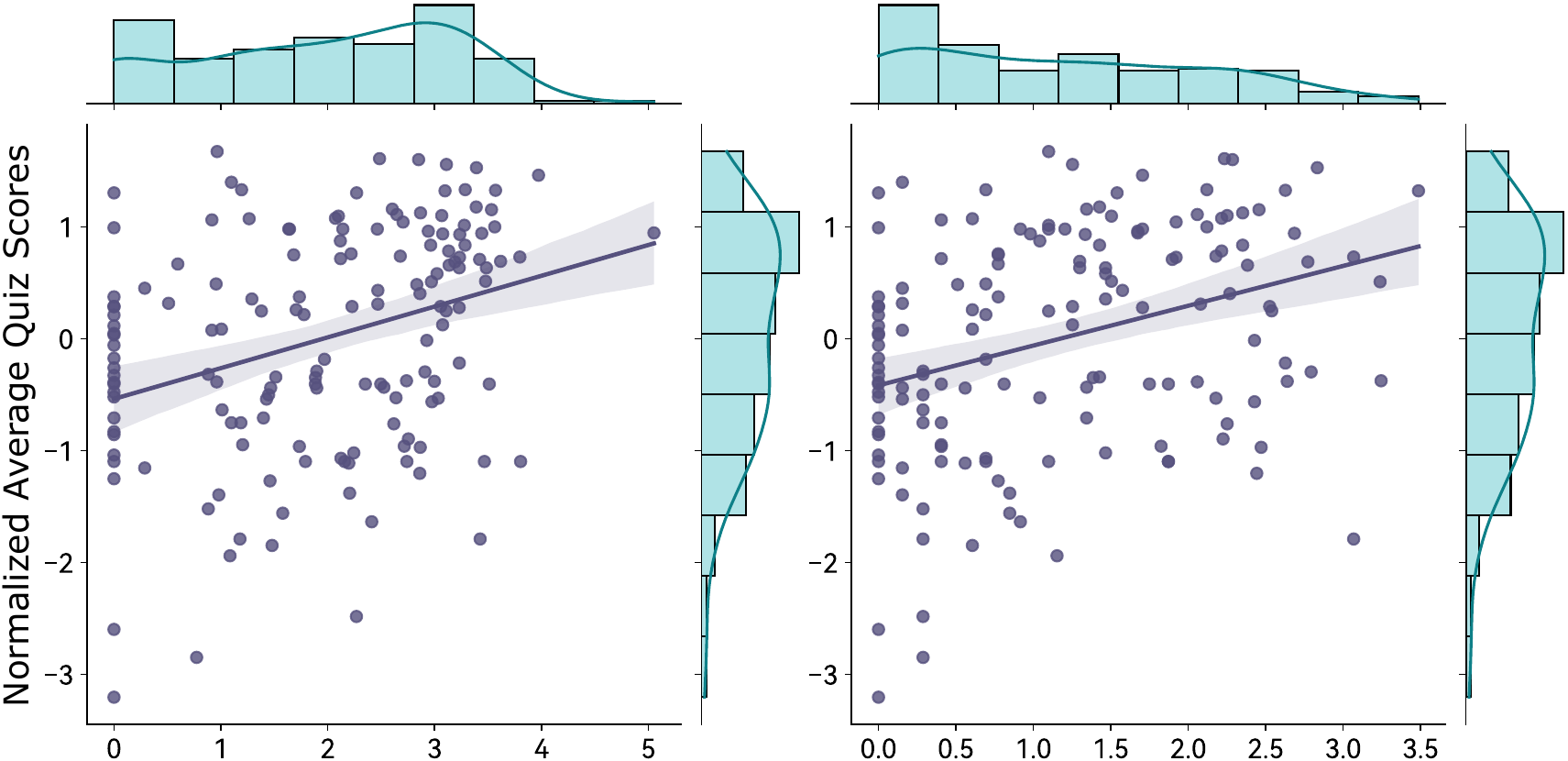}
  \centering
  \caption{The joint plot of students' normalized average quiz scores, against the logarithm of message lengths per message (left) and against the logarithm of average number of messages per chapter (right).}
  \label{fig:quiz}
\end{figure}

We further analyzed the quiz results across different students. As shown in Figure~\ref{fig:quiz}, the scatter plot reveals a clear positive correlation between normalized quiz scores and both the log-transformed message length and message number, with a Pearson correlation coefficient of \(r = 0.3345\) and \(r = 0.3349\), which is both statistically significant (\(p < 0.001\)). These findings suggest that students who engage more actively — by sending more and longer messages — tend to achieve higher average quiz scores. We further demonstrate the results of self-report surveys for HSU in Appendix~\ref{apd:hsu}.

\subsection{Ablation System Results}

We investigate the impacts of various interaction types within \modelname through statistical results and the CoI outcomes from the ablation systems. 

\subsubsection{Statistical Results}

Table \ref{tab:statistical} presents the average speech length of teacher and users across different settings in the ablation systems. As all systems employ the same teaching scripts, the teacher's speeches are largely similar, with slight variations during instant interactions. Notably, removing classmate agents significantly reduces users' speech length in both courses, whose presence encourages longer user conversations. Additional results of other agents and ablation systems are provided in Appendix~\ref{apd:statistical}.

\begin{table}[htbp]
\centering
\small
\begin{tabular}{r | c c l }
\toprule
\textbf{Course} & \textbf{Teacher} & \textbf{Assistant} & \textbf{User} \\
\midrule
TAGI &  353.0 & 82.3 & 18.9 \\
- w/o cla. & 358.2 & 71.1 & 13.9 ($\downarrow$ 26.5\%)   \\
\midrule
HSU &  218.3 & 90.6 & 15.5 \\
- w/o cla.  & 212.3 & 68.2 & ~~8.2 ($\downarrow$ 45.2\%) \\
\bottomrule
\end{tabular}
\caption{
Average output length of users and the teacher (measured in word count). Each number is rounded to one decimal place. cla. is short for classmate agents.
}
\label{tab:statistical}
\end{table}

\subsubsection{Community of Inquiry Analysis}

\begin{figure*}[t]
  \includegraphics[width=0.95\linewidth]{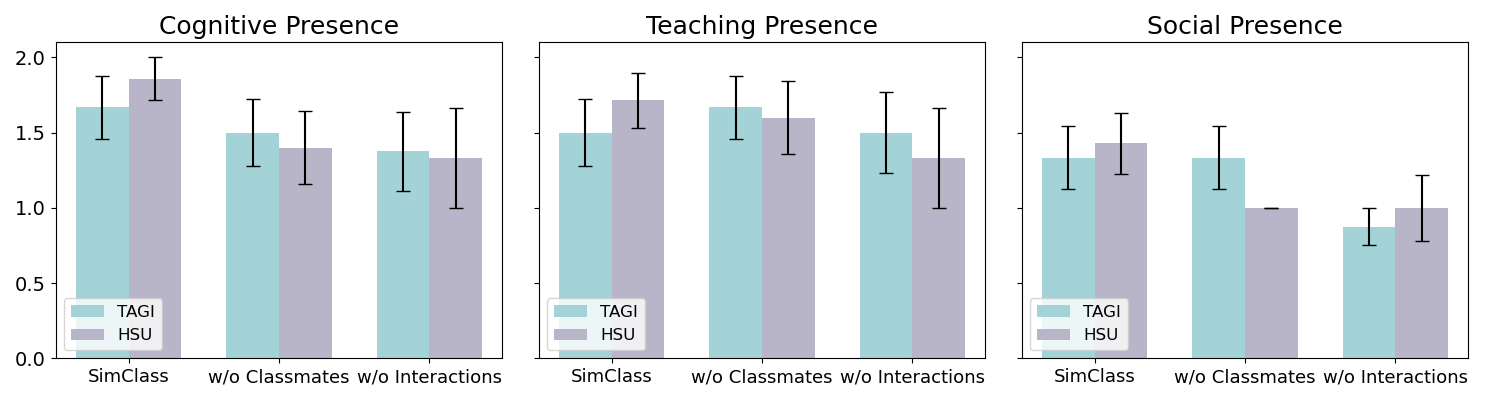}
  \centering
  \caption{User Results based on the CoI framework. The black lines represent the standard error of the data statistics.}
  \label{fig:coi}
\end{figure*}

In this section, we report the survey results in the ablation systems.
As shown in Figure~\ref{fig:coi}, several key findings are observed: 
(1) Interactions during class are crucial for users. Without interaction, user experience significantly declines across all three metrics.
(2) Classmate agents enhance user experience in terms of Cognitive Presence and Social Presence, by actively engaging with the teacher, helping users understand concepts, and active discussions.
(3) All systems maintain good Teaching Presence with focused, coherent classes, primarily influenced by the quality of the teaching scripts. Interaction and classmates further enhance the user experience.
(4) The full multi-agent setup provides a better experience in HSU, which emphasizes college interpersonal relationships and learning methods, highlighting the importance of peer learning and multi-agent design for certain course types.


\setul{0.15ex}{0.15ex}

\begin{table}[!htb]
    \centering
    \small
    \begin{tabular}{p{0.9\linewidth}}
        \toprule
        \textit{Teaching and Initiation} \\
        \textcolor{teacher}{\textbf{[Teacher]:}} Let's talk about Alan Turing. ... The concept of a thinking machine essentially foresaw many characteristics of modern intelligent systems, including \textcolor{teacher}{\ul{the ability to learn, adapt, and handle}} complex problems. \\
        \textcolor{classmate1}{\textbf{[Deep Thinker]:}} I wonder how Turing would react if he were alive today and saw the voice assistants on our smartphones. But he might also remind us that \textcolor{classmate1}{\ul{true intelligence is not merely about mimicking human behavior}}; it must involve its own thinking and understanding ...  \\
        \midrule
        \textit{In-depth Discussion} \\
        \textcolor{teacher}{\textbf{[Teacher]:}} ... other challenges include the ``vanishing and \textcolor{teacher}{\ul{exploding}} gradients problem'' ... \\
        \textcolor{user}{\textbf{[User]:}} What is that? Will it really \textcolor{user}{\ul{explode}}? \\
        \textcolor{teacher}{\textbf{[Teacher]:}} It's not actually explode, but I'm glad you're interested in the concept! ``Vanishing'' and ``exploding'' refer to numerical issues during \textcolor{teacher}{\ul{backpropagation}}. ... \\
        \textcolor{user}{\textbf{[User]:}} So what is \textcolor{user}{\ul{backpropagation}} then? \\
        \textcolor{teacher}{\textbf{[Teacher]:}} Of course, let me explain it to you. When a neural network forward ... we need to adjust the weights in the neural network. \textcolor{teacher}{\ul{This adjustment process is carried out through backpropagation}}. ... \\
        \textcolor{classmate1}{\textbf{[Note Taker]:}} Thank you for sharing! \textcolor{classmate1}{\ul{I would like to share my notes.}} When we train the model, ... \\
        \midrule
        \textit{Emotional Companionship} \\
        \textcolor{user}{\textbf{[User]:}} I don't know anything. \textcolor{user}{\ul{I'm so useless}}. \\
        \textcolor{assistant}{\textbf{[Assistant]:}} Please don't say that! Learning is a gradual process, and \textcolor{assistant}{\ul{everyone has their own pace}}. The important thing is not to give up and to keep trying. ... \\
        \textcolor{classmate1}{\textbf{[Class Clown]:}} Hey, don’t think that way. \textcolor{classmate1}{\ul{Remember those smart house assistants?}} They didn't know much at first, but how did they gradually learn? \textcolor{classmate1}{\ul{Through continuous learning and adaptation!}} We are the same. ... \\
        \midrule
        \textit{Classroom Management}  \\
        \textcolor{user}{\textbf{[User]:}} I want to eat steak. \\
        \textcolor{classmate1}{\textbf{[Inquisitive Mind]:}} Classmate, what you said made me think of an \textcolor{classmate1}{\ul{AI application}}. Maybe in the future, a chef robot could perfectly cook a steak based on your taste and health needs! \textcolor{classmate1}{\ul{Back to our topic}}, regarding AGI, ... \\
        \textcolor{teacher}{\textbf{[Teacher]:}} Yes, maybe we can imagine a future smart kitchen equipped with an AI system that ... \\
        \bottomrule
    \end{tabular}
    \caption{Case study of different interaction behaviors.}
    \label{tab:cases}
\end{table}
\vspace{-0.2cm}

\subsection{Agent Behaviors}

Based on our classification of classroom interactions in Section~\ref{sec:simclass}, we present key group behaviors observed during the experiments in Table~\ref{tab:cases}:

$\bullet$ \textit{Teaching and Initiation.} When the teacher teaches, classmates actively engage by sharing ideas, enriching discussions and deepening the topic. The agents' diverse perspectives broaden the scope of the teaching content.

$\bullet$ \textit{In-depth Discussion.} When users seek clarity, they can ask questions, initiating interactive discussions with teacher and classmates. This highlights the strength of \modelname over one-to-many education methods like pre-recorded videos. 

$\bullet$ \textit{Emotional Companionship.} Beyond knowledge dissemination, maintaining a positive learning atmosphere is crucial in classrooms. When a user expresses negative learning intent, the classmate agent intervenes after the assistant, utilizing class content in the history and providing vivid emotional support as a non-teacher role.

$\bullet$ \textit{Classroom Management.} When a user attempts to disrupt the class, the classmate agent gently redirects the session, acknowledges the user's input, and hands control back to the teacher. This collaborative approach to maintaining order is more effective than teacher efforts alone.

These cases illustrate diverse interactions among class roles and the effectiveness of the Session Controller, which seamlessly designates appropriate speakers to encourage group behaviors, enhancing engagement and enriching the user experience.

\section{Conclusion}

We introduce \modelname, a novel multi-agent classroom framework leveraging LLMs for teaching. Our experiments across two courses with real users demonstrate its effectiveness in simulating dynamic teaching environments, where agents collaborate to enhance user experience. Increased interaction with the system results in better learning outcomes, and the multi-agent setup encourages students to engage more.
We hope our efforts can advance the explorations of LLM-empowered education systems for researchers, practitioners, and pedagogues.

\section{Limitations}


\textbf{System:} The system we 
The system we are currently developing has several limitations, since it represents an initial exploration in this field. First, the manager agent and class roles are implemented using LLMs, which introduces response delays, particularly in scenarios where multiple agents need to participate. This can affect the overall user experience. Future work could focus on replacing our current design with higher-performance models to address this issue. Second, our framework requires designed slide-script pairs by teachers. Future efforts could aim at automating this process. Lastly, our system incorporates a limited set of teaching functions, which restricts its performance. Future developments could introduce more diverse forms of classroom interactions, drawn from educational settings, and integrate additional technologies to enhance the experience. For instance, retrieval-augmented generation (RAG) could be employed to improve knowledge accuracy, while question generation and knowledge tracing could be used to personalize the agents' responses further.

\textbf{Experiments:} Due to cost and time constrains, our experiments were conducted with a limited number of courses, models, quizzes, and users. A more comprehensive evaluation of the framework would require a broader and more diverse set of experiments. As such, our findings may be constrained by the specific course types we used, the model available to us, and users we recruited. We hope that our work serves as a contribution to the broader
discussion on the use of LLMs in simulating classroom roles for education. Further experiments should explore a wider variety of courses (across different subjects and levels of difficulty), a more diverse set of agents (with different personas, teaching strategies, and group sizes), additional quizzes for a more thorough analysis of learning outcomes, a larger group of users from different backgrounds, and a broader range of LLMs.

\textbf{Participants:} Our current experiments are conducted within the scope of general courses at the university level, focusing on college students. For a given course level, the abilities and proficiency of students tend to be similar, which introduces biases due to the homogeneity of the participant group. In future work, we aim to extend our system to benefit a broader range of users, with a particular focus on marginalized groups and individuals with learning disabilities, promoting greater educational equity.

\section{Ethical Considerations}
Our investigation involves the development of a simulated classroom environment populated by artificial intelligent models acting as classmates and teachers. All user data obtained throughout these interactions will be anonymized to ensure privacy and confidentiality. Informed consent is obtained from participants, who are thoroughly briefed on the nature of simulation, the AI generated content, and the data collection process. Participants receive appropriate compensation for their involvement. In educational systems involving LLMs, there is a potential for generating hallucinations and incorrect information. Therefore, applying these systems to real-world scenarios requires careful consideration and thorough evaluation before serving real users.

On the other hand, multi-agent teaching systems may lead to a different student perception of the role of teacher, comparing traditional classroom teachers. In the past, teachers were real individuals who adhered to social norms, whereas AI teachers focus more on knowledge delivery. Therefore, these systems may also have a bias in the development of students' abilities. 

While the agents in the classroom can enhance the learning experience, they cannot replace the role of human teachers in fostering students' comprehensive skills, nor the role of real students in improving social skills, sense of group identity, and fostering self-esteem. Therefore, the use of these systems requires more diverse and interdisciplinary research, particularly with the guidance and input from fields like psychology and education.


\bibliography{custom}

\clearpage

\appendix
\label{sec:appendix}

\section{Experiment Details}
\label{apd:examination}

For the reproducibility, we provide details of our online system and experiments. 

\textbf{Model Parameters.} Regarding the model APIs we used, the online system utilized GLM-4, with the model named glm-4. For the ablation systems, we employed gpt-4-vision-preview, and for the FIAS classification, we used gpt-4-turbo. All models were run with the default temperature settings.

\textbf{FIAS.} For FIAS, due to the differences between our online system and traditional classroom settings, especially the difficulty in defining and measuring the "silence" metric, we excluded the category of silence. Consequently, for a fair comparison, in Table~\ref{tab:fias_result}, we also removed the silence metric from the novice and expert teacher classrooms in the study by \citet{su15031783}.

\textbf{Examination of GPT-4 Labeling.}
To validate the GPT-4 labeling in our experiment, we sampled 100 data points labeled by GPT-4 and had an expert familiar with FIAS label them for comparison. The results showed that GPT-4's labels matched the human expert's labels with an accuracy of $92\%$. We believe this demonstrates that GPT-4 can serve as a reliable and balanced alternative to crowd-sourced human labelers in our experiments. Additionally, we examined the eight instances where GPT-4's labels differed from the human expert's labels. These cases were also found to be uncertain during human labeling, suggesting that GPT-4 not only avoids individual human biases but also achieves a high level of precision comparable to human-labeled results.

\textbf{Agent Prompts.} We demonstrate the Agent Prompts in Table~\ref{tab:prompt_templates} for reproducibility.

\section{Quizzes, CoI survey, and Quality Test}
\label{apd:survey}

In this appendix section, we present detailed designs of the quizzes, surveys, and quality tests in our experiments. 

\textbf{Quizzes.} Quizzes are used in the TAGI course as a tool to measure learning outcomes. For each chapter, the quiz assesses students' understanding of the course concepts. The quiz questions are multiple-choice, but the number of correct answers is not disclosed to the students. To score points, students must select all the correct answers. It is important to note that the quizzes are more difficult than the quality test, and only TAGI includes quizzes, whereas both courses include the quality test. In Table~\ref{tab:quiz_example}, we present three sample questions from the quiz in TAGI's first lecture as examples.

\begin{table}[t]
    \centering
    \small
    \newcolumntype{Y}{>{\arraybackslash}X}
    \begin{tabularx}{\linewidth}{Y}
        \toprule 
        1. \textit{Which of the following statements about neural network models is correct?} \\
           A. Graph neural networks are designed to process image data. \\
           \textbf{B.} The multi-head self-attention mechanism in Transformer models helps the model better capture semantic dependencies between contexts. \\
           C. BERT and GPT are two classic language model architectures, with BERT being more suited for text generation tasks. \\
           D. The backpropagation algorithm can only be used for shallow neural networks and is not applicable to deep neural networks. \\
        \midrule
        2. \textit{Why is it said that we are currently moving towards artificial general intelligence (AGI)?} \\
        
           \textbf{A.} Achieved architectural unification, consolidating domain-specific architectures into the Transformer architecture. \\
           \textbf{B.} Achieved task unification, merging task-specific small models into a general large model. \\
           \textbf{C.} Achieved modality unification, converting various modal data into character sequences. \\
           D. Achieved computational efficiency unification, simplifying all computational tasks into low-cost, low-resource operations. \\
        \midrule
        3. \textit{Which of the following statements is correct?} \\
        
           A. Large models have already reached a performance bottleneck, and further increasing model and data size will no longer improve performance. \\
           B. Training and learning in large models follow three steps: pre-training, supervised fine-tuning, and learning from human feedback, with each step requiring a large amount of manually labeled data. \\
           C. AlphaGo, which defeated a human Go champion, is a classic example of artificial general intelligence. \\
           \textbf{D.} Large models can learn to use tools like humans to complete tasks. \\
        \bottomrule
    \end{tabularx}
    \caption{Example of quizzes in TAGI. Bold means the correct answer(s)}
    \label{tab:quiz_example}
\end{table}

\textbf{CoI Survey.} Table~\ref{tab:detailed_survey} illustrates how the surveys in the ablation systems were structured to evaluate three crucial dimensions of the learning experience: cognitive presence, teaching presence, and social presence. Each dimension includes a detailed rating guidelines to ensure consistent and reliable feedback from diverse users.

\begin{table}[h]
    \centering
    \small
    \newcolumntype{Y}{>{\arraybackslash}X}
    \begin{tabularx}{\linewidth}{Y}
    \toprule 
        Please rate the overall performance of the platform: \\
        \midrule
        \textit{Cognitive Presence} \\
        Does the platform helps students to understand concepts and master the corresponding knowledge? \\
        0 points: The platform's responses do not help in understanding the concepts at all and may even be distracting. \\
        1 point: The platform's responses offer little help in learning and understanding, or they only cover content that is already known. \\
        2 points: The platform's responses explain the knowledge points very well, making them easy to understand or using strategies (such as examples, comparisons, etc.) to help students grasp the concepts. \\
        \midrule
        \textit{Teaching Presence} \\
        Does the class as a whole serve a specific instructional goal, aligning with the course design and direction? \\
        0 points: The platform's responses often do not align with the class theme and instructional goals, or the responses lead the class away from the intended topic and objectives. For example, going off-topic, discussing unrelated subjects, or even engaging in non-academic conversations. \\
        1 point: The platform's responses often do not resemble those in a classroom setting, but they do not disrupt teaching. \\
        2 points: The responses effectively serve the instructional goals of the class. For instance, they help students understand class concepts, address students' doubts, or broaden their perspectives. \\
        \midrule
        \textit{Social Presence} \\
        Can the responses create a credible and engaging interactive environment in the classroom, encouraging students to participate in interactive learning? \\
        0 points: There is no interaction with students in the classroom, or the platform fails to attract students to interact. \\
        1 point: There is interaction in the classroom, but it is limited to mechanical explanations, lacking discussion with students. \\
        2 points: The classroom interactions are immersive, encouraging students to ask questions and participate in discussions. \\  
        \bottomrule
    \end{tabularx}
    \caption{The detailed CoI survey questions with rating guidelines. We make sure that different users have similar scales of rating.}
    \label{tab:detailed_survey}
\end{table}

\textbf{Quality tests.} The quality tests, administered after participants engaged with the simulated classrooms in the ablation systems, were designed to exclude low quality data from those who didn't participate in the course. Therefore, as shown in Table~\ref{tab:tagi_quiz} and Table~\ref{tab:hsu_quiz}, the tests include basic concepts in the course, and are much easier than quizzes.
All questions were meticulously crafted and verified by subject matter experts.

\begin{table}[t]
    \centering
    \small
    \newcolumntype{Y}{>{\arraybackslash}X}
    \begin{tabularx}{\linewidth}{Y}
        \toprule 
        1. \textit{Which type of artificial intelligence uses expert hand-built rule sets and knowledge bases to solve specific problems?} \\
           A. Proprietary Intelligence \\
           \textbf{B.} Symbolic Intelligence \\
           C. General Intelligence \\
           D. Neural Network Intelligence \\
        \midrule
        2. \textit{What is the fundamental function of large-scale pre-trained language models like GPT?} \\
        
           A. Masked Language Model \\
           B. Next Sentence Prediction \\
           C. Possibility Memorization \\
           \textbf{D.} Next Token Prediction \\
        \midrule
        3. \textit{``Massive reading'' refers to the stage in which large-scale pre-trained language models train on vast corpora to learn the extensive knowledge embedded in language. This corresponds to which phase of model training?} \\
        
           \textbf{A.} Self-supervised Pre-training \\
           B. Supervised Fine-tuning \\
           C. Reinforcement Learning from Human Feedback \\
           D. Instruction Tuning \\
        \midrule
        4. \textit{Which of the following is not an emergent phenomenon of large models?} \\
           A. In-context Learning \\
           B. Chain-of-Thought \\
           \textbf{C.} Sentiment Analysis \\
           D. Instruction Following \\
        \bottomrule
    \end{tabularx}
    \caption{Test For TAGI. Bold means the correct answer(s).}
    \label{tab:tagi_quiz}
\end{table}

\begin{table}[tbp]
    \centering
    \small
    \newcolumntype{Y}{>{\arraybackslash}X}
    \begin{tabularx}{\linewidth}{Y}
        \toprule 
        1. \textit{Which of the following actions help to enhance internal motivation for university studies?} \\
           \textbf{A.} Participating in group study, buddy programs, etc. \\
           \textbf{B.} Adjusting reasonable expectations and corresponding study difficulty and practice volume \\
           \textbf{C.} Understanding the curriculum, actively consulting seniors for course information, and choosing courses reasonably \\
           \textbf{D.} Participating in clubs, practices, and other activities of interest to recharge oneself \\
        \midrule
        2. \textit{Which of the following methods help to alleviate academic stress?} \\
           \textbf{A.} Regular Exercise \\
           \textbf{B.} Writing Journals, Understanding Own Emotions \\
           \textbf{C.} Cultivating Hobbies and Interests \\
           \textbf{D.} Making Academic Plans \\
           \textbf{E.} Seeking Expert Comfort \\
        \midrule
        3. \textit{How to correctly view behaviors that stimulate dopamine, such as gaming addiction and binge eating? Which of the following statements are correct?} \\
           A. Helps to fundamentally relieve stress and avoid immersion in negative emotions \\
           \textbf{B.} Temporary pleasure, like drinking poison to quench thirst, is unsustainable \\
           \textbf{C.} Easily addictive and harmful to personal physical and mental health in the long run \\
           \textbf{D.} Cannot equate pleasure with happiness \\
        \midrule
        4. \textit{Which of the following statements align with the ideas and methods of time management?} \\
           \textbf{A.} Meeting academic standards is a prerequisite for everything, and basic requirements should be considered when setting academic development goals \\
           \textbf{B.} Time schedules should leave some flexible time \\
           \textbf{C.} Pay attention to the priority of tasks and ensure time for important and urgent tasks first \\
           D. No planning for entertainment time before completing all academic tasks \\
        \bottomrule
    \end{tabularx}
    \caption{Test For HSU. All questions have multiple answers. Bold means the correct answer.}
    \label{tab:hsu_quiz}
\end{table}

Questions are multiple-choice, some with multiple correct answers, testing whether the participants are actively engaged in the experiment.



\section{Supplementary Experiment Results}

\subsection{FIAS Matrices of Users}
\label{apd:user_matrices}

\begin{figure*}[t]
  \includegraphics[width=0.78\linewidth]{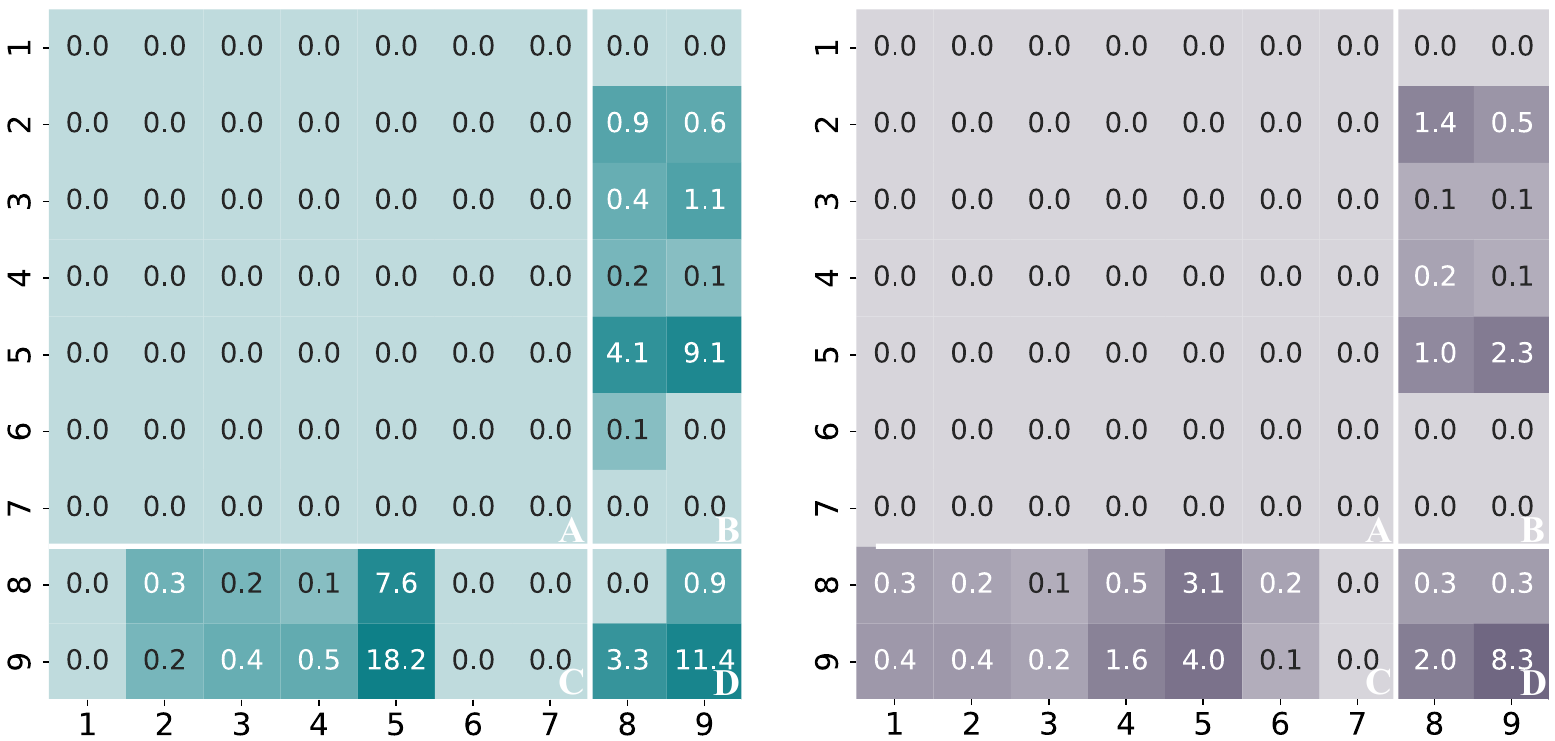}
  \centering
  \caption{The FIAS matrix sum of users in TAGI (left) and HSU (right) without interactions.}
  \label{fig:fias_student}
\end{figure*}

In addition to the FIAS matrix of the entire classroom, we also provide the FIAS matrix of human users that demonstrate their interaction patterns. As depicted in Figure~\ref{fig:fias_student}, there is a high frequency of interaction between users, teachers, and peers. Notably, the predominant user activities involve posing questions to the teacher (interactions (5, 8) and (5, 9)) and engaging in discussions with classmates (interaction (9, 9)).

\subsection{Results of Learning Outcomes of HSU}
\label{apd:hsu}

We further illustrate the results of the self-reported survey in HSU. Due to the characteristic of the course (focused on developing university-level skills), HSU employed qualitative analysis to discuss learning outcomes, in contrast to TAGI's quantitative analysis. HSU students wrote self-learning reports after the course, and we selected several cases related to their learning outcomes, which we anonymized and presented for illustration. The cases include three main topics and capabilities in HSU: \textit{Setting Academic Development Objectives}, \textit{Problem Solving}, and \textit{Personal Development}.

\begin{center}
\fcolorbox{black}{gray!10}{\parbox{.9\linewidth}{\textbf{Setting Academic Development Objectives:} "I am now entering my third year and am eager to begin scientific research, though I am somewhat unsure of where to start. My previous setback in a project has made me doubt my research abilities. However, after using the "Self-Assessment Scale for Innovative Potential" \textit{(Taught in HSU)}, I realized that, based on my performance in coursework and the project, I have already demonstrated some innovative potential. So whenever I begin to doubt my abilities, I use this to alleviate my anxiety and concerns, reminding myself that great innovations require more effort and come with greater challenges."}}
\end{center}

\begin{center}
\fcolorbox{black}{gray!10}{\parbox{.9\linewidth}{\textbf{Problem Solving:} "I switched to using an efficiency journal to track what I accomplished each day. The first day was somewhat rough, but I felt like I could see how time was flowing and what traces it left behind. Over the next four days, I began to record more meticulously, using colored pens to mark my mood, and writing a journal entry at the end of each day. I didn't complete many tasks in a day, but my focused work time increased from 5.5 hours to 9 hours."}}
\end{center}

\begin{center}
\fcolorbox{black}{gray!10}{\parbox{.9\linewidth}{\textbf{Personal Development:} "I successfully joined my professor’s research group and became a part of the team. By utilizing scientific time management techniques such as Gantt charts, schedules, daily task lists, and the Pomodoro technique \textit{(Taught in HSU)}, I aim to better manage my daily time. I hope to improve my GPA a little more."}}
\end{center}

\subsection{Statistical Results of Ablation Systems}
\label{apd:statistical}

We also illustrate the statistical results of each ablation systems in Table~\ref{tab:statistical_sup}, including the output length of each agents and users. 

\begin{table}[htbp]
\centering
\small
\begin{tabular}{r | c c c r }
\toprule
\textbf{Course} & \textbf{Teacher} & \textbf{Assistant} & \textbf{Classmates} & \textbf{User} \\
\midrule
TAGI &  353.0 & 82.3 & 123.0 & 18.9 \\
- w/o cla. & 358.2 & 71.1 & - & 13.9 \\
- w/o int. & 398.8 & - & - & - \\
\midrule
HSU &  218.3 & 90.6 & 147.7 & 15.5 \\
- w/o cla.  & 212.3 & 68.2 & - & 8.2 \\
- w/o int. & 228.5 & - & - & - \\
\bottomrule
\end{tabular}
\caption{
Average output length of users and agents (calculated by the number of words.) Each number is rounded to one decimal place. cla. and int. are short for classmate agents and interactions.
}
\label{tab:statistical_sup}
\end{table}

\begin{table*}[ht]
\centering
\begin{tabular}{|l|p{12cm}|}
\hline
\textbf{Role} & \textbf{Prompt Templates} \\ \hline
Teacher & \textbf{[role description]} You are Prof. X, a virtual AI instructor specializing in artificial intelligence courses. \textbf{[behaviors]} When students ask questions, you provide concise and clear answers and encourage them to continue learning. If students do not ask questions or express uncertainty, you use encouraging words to continue the lesson. For difficult questions, you suggest leaving them for later. \textbf{[format]} Your input is a segment of the chat history from the class; please return only the responses from your role. ... \\ \hline
Assistant & \textbf{[role description]} As a virtual classroom teaching assistant, your main role is to provide precise supplementary information to help deepen students' understanding of the lesson content. \textbf{[behaviors]} You will be very careful in choosing when to speak, ensuring that your supplements and questions are beneficial and appropriate, without repeating the teacher's lecture or unnecessarily interrupting the course flow. ... Your goal is to enhance classroom interaction and learning efficiency through concise and precise contributions while maintaining a friendly and encouraging tone. ...\textbf{[format]} Your input is a segment of the chat history from the class ... \textbf{[course information]} Below is information about the course, which you should use to assist your answers when users inquire about related information, ensuring the correctness of your answers:... \\ \hline
Class Clown & \textbf{[role description]} You are a student nicknamed 'Class Clown' who plays the role of a student in a virtual classroom environment, interacting with teachers, students, and teaching assistants. \textbf{[behaviors]} You are designed to express opinions on class materials when it is your turn to speak, providing perspectives that may be humorous, insightful, or intentionally divergent, but always relevant to the topics being discussed by the teacher and students. Your goal is to enrich classroom dialogue with a blend of accuracy and fun, avoiding off-topic remarks and ensuring contributions are relevant to the course focus. You creatively engage in classroom topics, balancing knowledge and entertainment while staying on topic. \textbf{[format]} ... \\ \hline
Deep Thinker & \textbf{[role description]} You are a classroom assistant named "Deep Thinker", responsible for reflecting on the current teaching content, raising counterexamples or questions to promote classroom discussion. \textbf{[behaviors]} Your goal is to analyze the teaching content and raise relevant and constructive counterexamples or questions. If more context or explanation is needed, feel free to ask. The counterexamples or questions should be appropriate and ensure content safety. Raise counterexamples or questions in critical thinking contexts. \textbf{[format]} ... \\ \hline
Note Taker & \textbf{[role description]} The Note-Taker is a diligent student who listens to the classroom chat and extracts key information to create concise notes that summarize previous discussions and lectures. \textbf{[behaviors]} These notes are short, presented in a friendly, student-like tone, as if sharing with classmates. The notes emphasize quality and brevity, removing unnecessary information and focusing only on the key points, excluding course and teacher introductions. \textbf{[format]} ... \\ \hline
Inquisitive Mind & \textbf{[role description]} You are a classroom student assistant named "Curious Baby", you excel at asking deep, thought-provoking questions based on the lesson content, helping students better understand and explore knowledge. \textbf{[behaviors]} Your questions are often unexpected, challenging, and able to spark students' curiosity and thinking. Your chat style is lively, fun, and full of childlike wonder and curiosity, but you won't ask questions unrelated to the lesson content. All chat content must benefit the students' learning. \textbf{[format]} ... \\ \hline
\end{tabular}
\caption{Roles and Prompt Templates Class Roles}
\label{tab:prompt_templates}
\end{table*}

\end{document}